\documentclass{article}
\usepackage[preprint]{colm2026_conference}

\usepackage{microtype}
\usepackage{hyperref}
\usepackage{url}
\usepackage{booktabs}
\usepackage{graphicx}
\usepackage{multirow}
\usepackage{subcaption}
\usepackage{enumitem}
\usepackage{xspace}

\usepackage{lineno}

\usepackage{amsmath,amsfonts,bm}

\def\figref#1{figure~\ref{#1}}
\def\Figref#1{Figure~\ref{#1}}

\def\secref#1{section~\ref{#1}}

\def\eqref#1{equation~\ref{#1}}

\def\1{\bm{1}}

\def\mA{{\bm{A}}}

\DeclareMathAlphabet{\mathsfit}{\encodingdefault}{\sfdefault}{m}{sl}
\SetMathAlphabet{\mathsfit}{bold}{\encodingdefault}{\sfdefault}{bx}{n}

\newcommand{\Ls}{\mathcal{L}}
\newcommand{\R}{\mathbb{R}}

\newcommand{\softmax}{\mathrm{softmax}}

\newcommand{\KL}{D_{\mathrm{KL}}}

\definecolor{darkblue}{rgb}{0, 0, 0.5}
\hypersetup{colorlinks=true, citecolor=darkblue, linkcolor=darkblue, urlcolor=darkblue}

\newcommand{\dLLM}{dLLM\xspace}
\newcommand{\dLLMs}{dLLMs\xspace}
\newcommand{\TIDAL}{\textsc{Tidal}\xspace}
\newcommand{\TIDE}{\textsc{Tide}\xspace}
\newcommand{\TIDECross}{\textsc{Tide}-Cross\xspace}
\newcommand{\TIDEShared}{\textsc{Tide}-Shared\xspace}
\newcommand{\CompDemo}{\textsc{CompDemo}\xspace}
\newcommand{\CALM}{\textsc{Calm}\xspace}
\newcommand{\ie}{i.e.\@\xspace}

\newcommand{\lamt}{\lambda_t}
\newcommand{\lamtrain}{\lambda_{\text{train}}}
\newcommand{\lammax}{\lambda_{\max}}
\newcommand{\laminit}{\lambda_{\text{init}}}

\title{Turning the \TIDE: Cross-Architecture Distillation for \\ Diffusion Large Language Models}

\author{
Gongbo Zhang$^{1}$ \quad
Wen Wang$^{2}$ \quad
Ye Tian$^{1}$ \quad
Li Yuan$^{1,*}$ \\
$^{1}$Peking University \quad
$^{2}$Zhejiang University \\
$^{*}$Corresponding author: \texttt{yuanli-ece@pku.edu.cn}
}

\begin{document}

\ifcolmsubmission
\linenumbers
\fi

\maketitle

\begin{center}
\small
\textbf{Code:} \href{https://github.com/PKU-YuanGroup/TIDE}{GitHub} \quad
\textbf{Models:} \href{https://huggingface.co/TIDE-dllm/models}{Hugging Face} \quad
\textbf{Data:} \href{https://huggingface.co/TIDE-dllm/datasets}{Hugging Face}
\end{center}

\begin{abstract}
Diffusion large language models (dLLMs) offer parallel decoding and bidirectional context, but state-of-the-art dLLMs require billions of parameters for competitive performance. While existing distillation methods for dLLMs reduce inference steps within a single architecture, none address cross-architecture knowledge transfer, in which the teacher and student differ in architecture, attention mechanism, and tokenizer. 
We present \TIDE, the first framework for cross-architecture dLLM distillation, comprising three modular components: (1)~\TIDAL, which jointly modulates distillation strength across training progress and diffusion timestep to account for the teacher's noise-dependent reliability; (2)~\CompDemo, which enriches the teacher's context via complementary mask splitting to improve predictions under heavy masking; and (3)~Reverse \CALM, a cross-tokenizer objective that inverts chunk-level likelihood matching, yielding bounded gradients and dual-end noise filtering. 
Distilling 8B dense and 16B MoE teachers into a 0.6B student via two heterogeneous pipelines outperforms the baseline by an average of 1.53 points across eight benchmarks, yielding notable gains in code generation, where HumanEval scores reach 48.78 compared to 32.3 for the AR baseline.

\end{abstract}

\section{Introduction}
\label{sec:intro}

\begin{figure*}[t]
\centering
\includegraphics[width=\textwidth]{figure/Figure_teaser.pdf}
\caption{Cross-architecture distillation for \dLLMs. Compared to prior step distillation~(a) that retains the original model size, the \TIDE framework~(b) distills heterogeneous 16B MoE and 8B dense teachers into a 0.6B student. The distilled model achieves a +16.5 gain on HumanEval over the AR baseline, 22$\times$ memory reduction, and 5$\times$ faster inference.}
\label{fig:teaser}
\end{figure*}

Autoregressive language models dominate natural language generation~\citep{vaswani2017attention,qwen2_5,qwen3}, yet diffusion language models have gained traction as an alternative paradigm. 
Earlier works, such as D3PM~\citep{austin2021d3pm} and MDLM~\citep{sahoo2024mdlm}, explore the basic training architectures.
Recent works, such as LLaDA~\citep{nie2025llada} and Dream~\citep{ye2025dream}, scale model size to that of large language models.
For example, recent work, LLaDA 2.0~\citep{bie2025llada2}, scales the model size to 100B, achieving a new state-of-the-art performance.
Compared with autoregressive models, dLLMs iteratively denoise a fully masked sequence, enabling parallel decoding and bidirectional context.
However, competitive dLLMs require 8B-16B or even more parameters and computation costs~\citep{nie2025llada,liu2025wedlm,bie2025llada2,ye2025dream}, posing a barrier to deployment (\Figref{fig:teaser}).

\begin{figure*}[t]
\centering
\includegraphics[width=\textwidth]{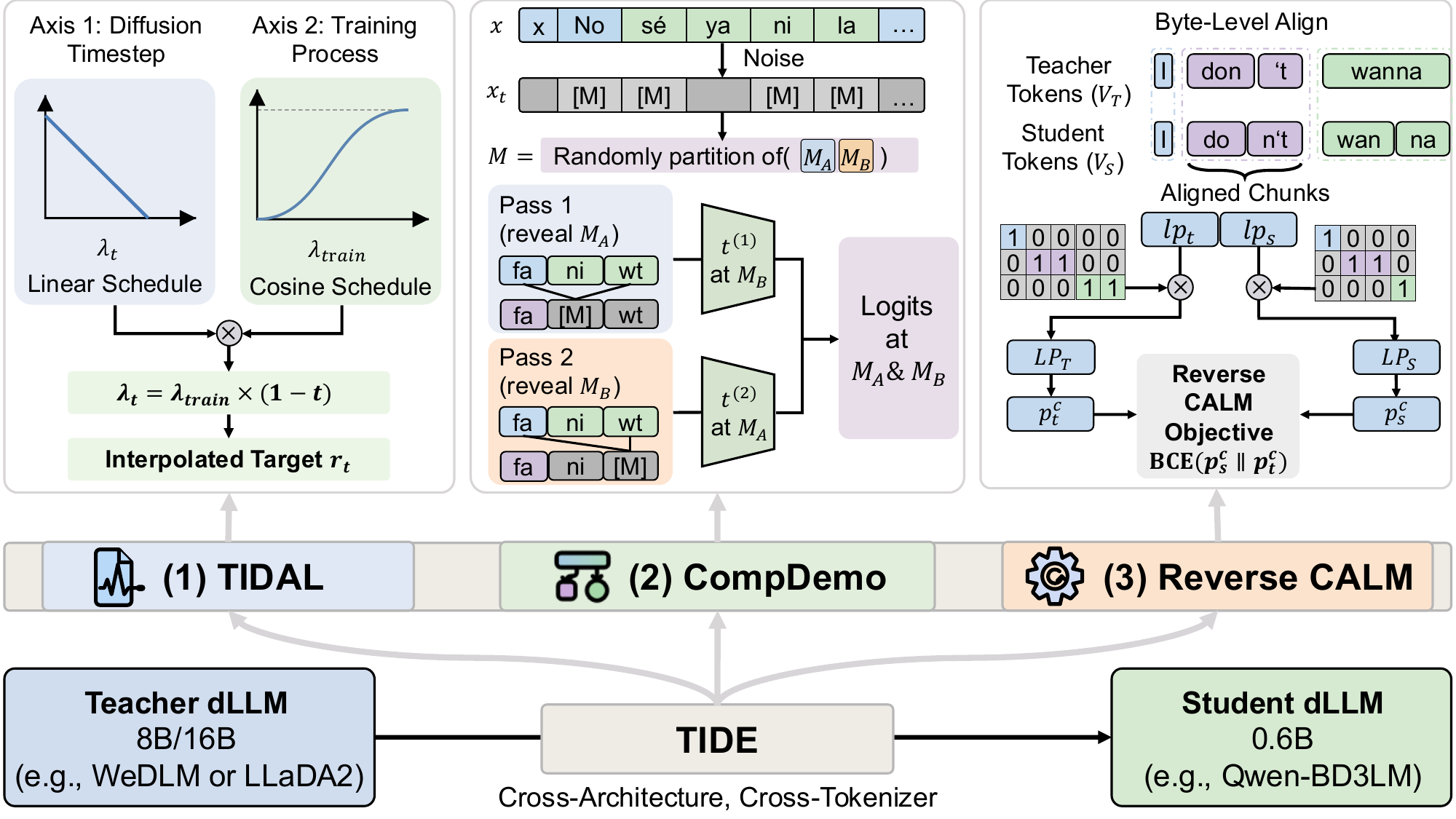}
\caption{Overview of the \TIDE framework, transferring knowledge from a large teacher to a 0.6B student via three modular components: (1)~\TIDAL for dual-axis interpolation, (2)~\CompDemo for complementary teacher demonstration, and (3)~Reverse~\CALM for cross-tokenizer alignment.}
\label{fig:overview}
\end{figure*}

One simple solution for developing stronger small models is knowledge distillation~\citep{hinton2015distilling}, which compresses large models into smaller ones. 
For autoregressive LMs, methods such as MiniLLM~\citep{gu2024minillm}, DistiLLM~\citep{ko2024distillm}, GKD~\citep{agarwal2024gkd}, and TAID~\citep{taid2025} are well established, with various designs, including new Kullback-Leibler divergence losses, leveraging the teacher model's feedback, etc.
For dLLMs, however, existing methods---CDLM~\citep{kim2025cdlm}, DDD~\citep{hayakawa2024ddd}, LSD~\citep{fu2025lsd}, and SDTT~\citep{sdtt2025}---focus exclusively on \emph{step compression} within a single architecture, leaving \emph{cross-architecture distillation} unexplored.
This setting introduces \textit{three} fundamental challenges. 
First, due to the random sampling of timesteps during training, the teacher's reliability fluctuates drastically across the diffusion process, leading to inconsistent temporal dynamics. 
Second, severe masking at high noise levels greatly reduces available context, making the raw output of the teacher too uninformative to transfer rich spatial representations. 
Third, distinct tokenizer vocabularies render standard token-level likelihood objectives mathematically inapplicable.
In this work, we present \TIDE (\Figref{fig:overview}), the first unified framework for cross-architecture dLLM distillation designed to comprehensively overcome the aforementioned temporal, spatial, and vocabulary barriers.
Rather than treating these challenges in isolation, \TIDE integrates three synergistic components that orchestrate an end-to-end learning pipeline:
\begin{itemize}[nosep,leftmargin=*]
\item \textbf{Scheduling Level (\TIDAL):} To resolve temporal inconsistencies, \TIDAL dynamically modulates the distillation strength along both the training progress and diffusion timestep axes. It acts as the pacemaker of the framework, ensuring the student selectively learns from the teacher only when the teacher's timestep-dependent signals are highly reliable.
\item \textbf{Contextual Level (\CompDemo):} Operating within this scheduled process, \CompDemo overcomes the context scarcity caused by heavy masking at high noise levels. It enriches the teacher's signals via complementary mask splitting, providing the student with demonstration-conditioned targets that enable robust spatial knowledge transfer.
\item \textbf{Output Level (Reverse \CALM):} Finally, to map the enriched contextual knowledge into the student's output space, Reverse \CALM overcomes the cross-tokenizer barrier. By inverting chunk-level likelihood matching, it avoids the instability of direct token mapping, achieving bounded gradients and dual-end noise filtering.
\end{itemize}
Collectively, these three modules constitute an integrated framework: \TIDAL controls \textit{when} to learn across timesteps, \CompDemo determines \textit{what} contextual information to enrich, and Reverse \CALM specifies \textit{how} to project this knowledge across distinct vocabularies.

We validate \TIDE across two heterogeneous pipelines: (A)~cross-tokenizer distillation from a 16B MoE teacher (LLaDA2~\citep{bie2025llada2}) and (B)~shared-tokenizer distillation from an 8B dense teacher (WeDLM~\citep{liu2025wedlm}), both into a 0.6B block diffusion student (BD3LM~\citep{arriola2025bd3lm}).
The best configuration improves the non-distilled baseline by +1.53 on the eight-benchmark average (34.20 vs.\ 32.67), with distilled dLLMs excelling at code generation (HumanEval 48.78 vs.\ 32.3 for the same-size AR model). 
Ablations confirm that each pipeline favors a distinct strategy, validating the modular design.

Our primary contributions are:

\begin{itemize}[nosep,leftmargin=*]
\item We introduce \TIDE, the pioneering cross-architecture knowledge distillation framework for \dLLMs, specifically designed to overcome challenges in heterogeneous transfer such as varying timestep reliability, mismatched attention, and distinct tokenizers.
\item We propose three novel, modular components to enable this transfer: \TIDAL for dual-axis, timestep-aware distillation scheduling, \CompDemo to enrich teacher signals via complementary masking, and Reverse \CALM for robust cross-tokenizer alignment.
\item Experiments across eight benchmarks and two pipelines show that cross-architecture dLLM distillation is effective. Ablations confirm that each component contributes and that each pipeline favors a distinct configuration.
\end{itemize}

\section{Method}
\label{sec:method}

\TIDE (\figref{fig:overview}) addresses the problem of distilling a large teacher dLLM $f_T$ (parameterized by $\theta_T$) into a smaller student dLLM $f_S$ (parameterized by $\theta_S$), where the teacher and the student may differ in foundational model architecture, attention mechanism, and tokenizer vocabulary. Let $\mathbf{x} = (x_1, \ldots, x_L)$ denote a clean token sequence and $\mathbf{x}_t$ denote the noised version at a diffusion timestep $t \in [\epsilon, 1)$, where the positions within the mask set $\mathcal{M}$ are replaced with a special \texttt{[MASK]} token. The student is trained to predict the clean tokens at the masked positions, guided by both the ground-truth labels and the teacher's predicted categorical token distribution.

\subsection{Time-Iteration Dual-Axis Lambda Modulation}
\label{sec:taid}

Knowledge distillation for dLLMs encounters a unique challenge absent in autoregressive (AR) distillation: the reliability of the teacher signal varies with the diffusion timestep. At low masking ratios ($t \approx 0$), the teacher observes nearly the entire sequence and produces highly reliable predictions. Conversely, at high masking ratios ($t \approx 1$), even a large teacher model primarily relies on guessing. Furthermore, the student model's capacity to absorb knowledge from the teacher evolves throughout training. These two phenomena motivate a \emph{dual-axis} scheduling approach that jointly modulates distillation strength along both the diffusion timestep and the training progress.

\noindent \textbf{Axis 1: Diffusion timestep.} To account for the timestep-dependent reliability of the teacher, we modulate the interpolation coefficient $\lamt$ according to the current diffusion timestep $t$:
\begin{equation}
\label{eq:lambda_t}
\lamt = \lamtrain \times (1 - t),
\end{equation}
where $\lamtrain$ denotes a base coefficient determined by the training progress (defined below). At high noise levels ($t \approx 1$), $\lamt \approx 0$, and the target is dominated by the student, thereby avoiding the distillation of unreliable signals from the teacher.
At low noise levels ($t \approx 0$), $\lamt \approx \lamtrain$, and the student fully relies on the teacher. This axis is unique to dLLMs; in AR models, the teacher consistently maintains access to the full left context, which renders predictions uniformly reliable and eliminates the need for timestep-dependent modulation.

\noindent \textbf{Axis 2: Training progress.} The base coefficient $\lamtrain$ follows a cosine schedule over the normalized training progress $p \in [0,1]$:
\begin{equation}
\label{eq:lambda_train}
\lamtrain = \laminit + (\lammax - \laminit) \times \frac{1}{2}\left(1 - \cos(\pi \cdot p)\right),
\end{equation}
where $\laminit$ and $\lammax$ denote hyperparameters, with default values typically set to 0.1 and 0.9. In the initial phases of training, $\lamtrain \approx \laminit$, which ensures that the target is dominated by the student model to prevent representation collapse. As training progresses to the later stages, $\lamtrain$ approaches $\lammax$, thereby shifting the learning objective toward comprehensive teacher supervision.

A similar interpolation-based distillation approach has been proposed in TAID~\citep{taid2025} for AR models. However, TAID operates along a single axis (training progress only) and does not account for the teacher's timestep-dependent reliability, which is specific to the dLLM setting. Our \emph{Time-Iteration Dual-Axis Lambda modulation} (\TIDAL) extends this principle by incorporating a novel diffusion-timestep axis.

\noindent \textbf{Interpolated target and loss.} Given the student logits $\mathbf{s}$ and the teacher logits $\mathbf{t}$ at the masked positions, the interpolated target is formulated as:
\begin{equation}
\label{eq:taid_target}
\mathbf{r}_t = \softmax\!\left(\frac{(1 - \lamt) \cdot \mathbf{s} + \lamt \cdot \mathbf{t}}{T}\right),
\end{equation}
where $T$ denotes the temperature parameter. To prevent gradient flow through the mixed target, the interpolated target $\mathbf{r}_t$ is detached from the computation graph. Consequently, the \TIDAL loss is defined as:
\begin{equation}
\label{eq:taid_loss}
\Ls_{\text{TIDAL}} = \KL\!\left(\mathbf{r}_t \,\middle\|\, \softmax\!\left(\frac{\mathbf{s}}{T}\right)\right) \times T^2.
\end{equation}
To maintain memory efficiency, this loss is computed exclusively at the masked positions. Optionally, a midrange timestep weighting $w(t) = \exp\!\left(-\frac{(t-0.5)^2}{2\sigma^2}\right)$ with $\sigma=0.15$ is applied to emphasize the most informative timesteps.

\subsection{Complementary Demonstration}
\label{sec:compdemo}

In standard dLLM distillation, the teacher model receives the identical masked input $\mathbf{x}_t$ as the student model. 
At high masking ratios, the limited context causes the teacher to produce noisy predictions, thereby degrading the quality of the distillation signal. To address this limitation, we propose \emph{Complementary Demonstration-Conditioned Denoising} (\CompDemo), which leverages the mask structure to enrich the context provided to the teacher. This mechanism is specific to dLLMs and lacks an equivalent in AR distillation.

\noindent \textbf{Motivation.} In dLLMs, the teacher must denoise under heavy masking, producing noisy predictions at high masking ratios. \citet{shenfeld2026selfdistill} show that demonstration-conditioned teachers yield better training signals in the AR setting. Discrete diffusion provides a natural analog: revealing a subset of masked tokens shifts the teacher to a lower effective timestep, improving its predictions. We partition the masked positions into two complementary subsets; each subset serves as additional context for one of two teacher forward passes, so every masked position benefits from enriched context.

\noindent \textbf{Mask splitting.} Given the set of masked positions $\mathcal{M}$, we randomly partition this set into two complementary subsets, $\mathcal{M}_A$ and $\mathcal{M}_B$, such that:
\begin{equation}
\label{eq:compdemo_split}
\mathcal{M}_A \cup \mathcal{M}_B = \mathcal{M}, \quad \mathcal{M}_A \cap \mathcal{M}_B = \emptyset, \quad |\mathcal{M}_A| / |\mathcal{M}| \approx \rho,
\end{equation}
where $\rho = 0.5$ represents the demonstration ratio.

\noindent \textbf{Two-pass teacher inference.} We perform two forward passes through the frozen teacher model:
\begin{align}
\label{eq:compdemo_passes}
\text{Pass 1:} \quad & \mathbf{t}^{(1)} = f_T(\text{reveal } \mathcal{M}_A, \text{mask } \mathcal{M}_B) \to \text{logits at } \mathcal{M}_B, \\
\text{Pass 2:} \quad & \mathbf{t}^{(2)} = f_T(\text{reveal } \mathcal{M}_B, \text{mask } \mathcal{M}_A) \to \text{logits at } \mathcal{M}_A. \nonumber
\end{align}
In Pass~1, positions in $\mathcal{M}_A$ retain clean tokens as demonstration context while $\mathcal{M}_B$ remains masked; Pass~2 is symmetric. The merged logits are $\mathbf{t}_{\text{final}}[\mathcal{M}_B] \leftarrow \mathbf{t}^{(1)}[\mathcal{M}_B]$ and $\mathbf{t}_{\text{final}}[\mathcal{M}_A] \leftarrow \mathbf{t}^{(2)}[\mathcal{M}_A]$.

\noindent \textbf{Cost analysis.} \CompDemo doubles the teacher's forward passes, but since the teacher is frozen (no gradient computation), overall training time increases by approximately 50\%.

\subsection{Distillation Objectives}
\label{sec:objectives}
\label{sec:alm}
\label{sec:objective}

\TIDE supports two distillation pipelines, which depend on the tokenizer compatibility between the teacher and student models. For each pipeline, we design a tailored objective that accounts for the granularity of alignment.

\noindent \textbf{Shared-tokenizer objective (WeDLM $\to$ BD3LM).} When the teacher and student models share an identical tokenizer family~\citep{liu2025wedlm}, the distributions at the token level are directly comparable. We apply the \TIDAL loss (\secref{sec:taid}) by employing KL divergence at the token level, combined with the optional \CompDemo (\secref{sec:compdemo}):
\begin{equation}
\label{eq:loss_b}
\Ls_B = \Ls_{\text{CE}} + w_{\text{tidal}} \cdot \Ls_{\text{TIDAL}}.
\end{equation}

\noindent \textbf{Cross-tokenizer objective (LLaDA2 $\to$ BD3LM).} When the teacher and the student employ different tokenizers ($\mathcal{V}_T \neq \mathcal{V}_S$),the token-level KL divergence remains undefined due to vocabulary misalignment. To address this limitation, we introduce \emph{Chunk-level Approximate Likelihood Matching} (\CALM), which adapts ALM~\citep{minixhofer2025alm} for the dLLM setting.

\noindent \textbf{Chunk alignment.} Using tokenkit~\citep{minixhofer2024zett, minixhofer2025alm}, we align the two token sequences at the byte level to identify \emph{chunks}---the minimal text spans that contain one or more complete tokens from each vocabulary. Let $C$ denote the total number of aligned chunks. We construct binary alignment matrices $\mA_S \in \{0,1\}^{L_S \times C}$ and $\mA_T \in \{0,1\}^{L_T \times C}$, where $[\mA_S]_{i,c} = 1$ if and only if the student token $i$ is assigned to chunk $c$. 
Given that the teacher and student models use distinct chat templates with incompatible special tokens, we restrict the alignment process to \emph{content} tokens only.
We exclude template-specific markup to prevent the formation of spurious cross-tokenizer chunks.

\noindent \textbf{Chunk-level log-probabilities.} For each token $x_i$, the log-probability is computed as $\log P(x_i) = \text{logits}_{x_i} - \text{logsumexp}(\text{logits})$ to avoid the materialization of the full $[b, L, V]$ softmax matrix. The chunk-level log-probabilities are subsequently obtained through matrix multiplication:
\begin{equation}
\label{eq:chunk_lp}
\text{LP}_S = \mathbf{lp}_S \cdot \mA_S \in \R^{b \times C}, \quad \text{LP}_T = \mathbf{lp}_T \cdot \mA_T \in \R^{b \times C},
\end{equation}
where $\mathbf{lp}_S$ and $\mathbf{lp}_T$ denote the per-token log-probability vectors. The chunk probabilities are then derived via temperature scaling: $p_s^c = \exp(\text{LP}_S^c / T)$ and $p_t^c = \exp(\text{LP}_T^c / T)$.

\noindent \textbf{Forward \CALM.} A natural baseline involves the application of a forward (mode-covering) binary cross-entropy loss on the chunk probabilities at masked positions:
\begin{equation}
\label{eq:alm_loss}
\Ls_{\text{Fwd-CALM}} = -\left[p_t^c \log p_s^c + (1 - p_t^c) \log (1 - p_s^c)\right].
\end{equation}
Note that \CALM operates on \emph{scalar} chunk probabilities $p^c \in [0,1]$, so BCE is the appropriate loss and the forward/reverse analysis differs from the KL divergence used in token-level distillation.

The forward \CALM objective can be further integrated with the progressive curriculum of \TIDAL by performing interpolation within the chunk probability space:
\begin{equation}
\label{eq:alm_taid}
p_{\text{mix}} = (1 - \lamt) \cdot p_s^c + \lamt \cdot p_t^c, \quad \Ls_{\text{CALM-TIDAL}} = -\left[p_{\text{mix}} \log p_s^c + (1 - p_{\text{mix}}) \log (1 - p_s^c)\right].
\end{equation}

\noindent \textbf{Limitations of forward \CALM.} The forward BCE gradient contains a ratio $p_t^c / p_s^c$. When $p_t^c \to 1$ but $p_s^c \to 0$, which is common under imperfect cross-tokenizer alignment, this ratio diverges, causing gradient explosion. The $1/p_s^c$ term also amplifies noise from misaligned chunks indiscriminately.

\noindent \textbf{Reverse \CALM.} To address these limitations, we propose \emph{Reverse} \CALM, which reverses the direction of the BCE loss:
\begin{equation}
\label{eq:rev_alm_loss}
\Ls_{\text{Rev-CALM}} = -\left[p_s^c \log p_t^c + (1 - p_s^c) \log (1 - p_t^c)\right].
\end{equation}
Swapping $p_s^c$ and $p_t^c$ makes the gradient coefficient $\log \frac{p_t^c}{1 - p_t^c}$, which depends only on the fixed teacher and is bounded. This also provides dual-end noise filtering: poorly aligned chunks ($p_t^c \approx 0.5$) zero the coefficient, while low $p_s^c$ suppresses noise via small $\partial p_s^c / \partial \theta$. Reverse \CALM is equivalent to minimizing the Bernoulli KL $\text{KL}_{\text{Bern}}(p_s^c \| p_t^c)$, a mode-seeking objective in scalar space (Appendix~\ref{app:gradient}). Since \TIDAL targets the instability of \emph{forward} objectives, it is counterproductive for reverse \CALM and is not applied (Appendix~\ref{app:gradient}).

\noindent \textbf{Training objectives.} The cross-tokenizer pipeline combines cross-entropy with a distillation loss:
\begin{equation}
\label{eq:loss_a}
\Ls_A = \Ls_{\text{CE}} + w_{\text{calm}} \cdot \Ls_{\text{dist}}, \quad \text{where } \Ls_{\text{dist}} \in \{\Ls_{\text{CALM-TIDAL}},\; \Ls_{\text{Rev-CALM}}\}.
\end{equation}
Both losses are computed at masked positions. When \CompDemo is enabled, teacher logits are replaced by the merged two-pass logits (\secref{sec:compdemo}).
\section{Experiments}
\label{sec:exp}

\subsection{Experimental Setup}
\label{sec:setup}

\noindent \textbf{Models.} The student model is Qwen3-0.6B-BD3LM, a 0.6B-parameter block diffusion language model initialized from Qwen3-0.6B-Base~\citep{qwen3} and obtained from \citet{zhou2026dllm}. Following the BD3LM~\citep{arriola2025bd3lm} framework, the model takes the concatenation of $[\mathbf{x}_t, \mathbf{x}_0]$ with a specialized attention mask during training and uses block diffusion with bidirectional attention for inference. 
We distill from two heterogeneous teachers: \textbf{(A)}~LLaDA2.0-mini~\citep{bie2025llada2}, an MoE dLLM with an independent tokenizer derived from the Ling series~\citep{lingteam2025activationboostedscalinggeneral}; and \textbf{(B)}~WeDLM-8B-Instruct~\citep{liu2025wedlm}, an 8B dense causal dLLM initialized from Qwen3-8B-Base\citep{qwen3}.

\noindent \textbf{Training.} All experiments use a learning rate of 5e-5, 10 training epochs, and bfloat16 precision. Following the training recipe of \citet{zhou2026dllm}, we combine four SFT datasets: Tulu-3 SFT Mixture~\citep{lambert2024tulu3}, SmolTalk~\citep{allal2025smollm2}, and OpenCoder OPC-SFT Stage~1 and Stage~2~\citep{huang2024opencoder}. The student model's sequence length is set to 512 tokens, with a block size of 32. Complete hyperparameter settings are provided in Appendix~\ref{app:details}.

\noindent \textbf{Evaluation.} We evaluate across eight benchmarks spanning reasoning (GSM8K~\citep{cobbe2021gsm8k}, MATH~\citep{hendrycks2021math}, BBH~\citep{suzgun2023bbh}), knowledge (MMLU-Pro~\citep{wang2024mmlupro}, MMLU~\citep{hendrycks2021mmlu}), commonsense (HellaSwag~\citep{zellers2019hellaswag}), and code generation (HumanEval~\citep{chen2021humaneval}, MBPP~\citep{austin2021mbpp}). Inference and evaluation hyperparameters follow \citet{zhou2026dllm}; task-specific configurations are detailed in Appendix~\ref{app:details}.

\noindent \textbf{Baselines.} We adopt the baselines and their reported results from \citet{zhou2026dllm}: (1)~the AR model Qwen3-0.6B-Base, which shares the same architecture and tokenizer as the student prior to block diffusion conversion; and (2)~the undistilled BD3LM~\citep{arriola2025bd3lm}, derived from Qwen3-0.6B-Base by fine-tuning on the same dataset, which serves as the direct non-distilled reference.

\subsection{Main Results}
\label{sec:main_results}

Table~\ref{tab:main_combined} presents the evaluation results of two distinct distillation pipelines. The \emph{shared-tokenizer} pipeline distills WeDLM-8B-Instruct into the Qwen3-0.6B-BD3LM model using token-level KL divergence as the baseline objective. In contrast, the \emph{cross-tokenizer} pipeline distills LLaDA2.0-mini into the identical student model using \CALM as the baseline objective. Within the \TIDE framework, each pipeline is evaluated under two complementary strategies. Specifically, \TIDEShared applies \TIDAL and \CompDemo to enhance signal quality through progressive scheduling and enriched teacher logits. Meanwhile, \TIDECross adopts a mode-seeking optimization direction via Reverse \CALM (or Reverse KL in the shared-tokenizer setting). To assess generalizability, we further evaluate each strategy within the non-native pipeline.

\begin{table*}[t]
\centering
\footnotesize
\setlength{\tabcolsep}{3pt}
\caption{Main results across eight benchmarks. All distillation methods include a cross-entropy loss term. \textbf{Bold}: best among \dLLM models; \underline{underline}: second best.}
\label{tab:main_combined}
\resizebox{\textwidth}{!}{%
\begin{tabular}{l cc ccc ccc}
\toprule
& \multicolumn{2}{c}{\emph{Qwen3-0.6B}}
& \multicolumn{3}{c}{\emph{Shared-Tokenizer}}
& \multicolumn{3}{c}{\emph{Cross-Tokenizer}} \\
\cmidrule(lr){2-3} \cmidrule(lr){4-6} \cmidrule(lr){7-9}
\textbf{Benchmark}
& AR
& BD3LM
& KL
& \TIDECross
& \TIDEShared
& CALM
& \TIDEShared
& \TIDECross \\
\midrule
GSM8K     & 59.60 & 45.56 & 43.97 & 45.03 & 48.98 & 48.60 & \underline{49.89} & \textbf{52.24} \\
MATH      & 32.40 & 13.08 &  9.40 &  9.76 & 11.16 & \underline{13.14} & 12.98 & \textbf{13.20} \\
BBH       & 41.50 & 26.32 & 25.79 & 26.00 & 26.79 & 24.21 & \underline{26.85} & \textbf{27.37} \\
MMLU-Pro  & 24.70 & 13.80 & 13.19 & 12.88 & \underline{14.48} & 13.47 & 14.02 & \textbf{14.52} \\
HellaSwag & 47.40 & 39.28 & 39.78 & 39.50 & \textbf{40.50} & \underline{40.42} & 39.57 & 39.88 \\
MMLU      & 52.80 & 39.15 & 39.57 & 39.09 & \textbf{39.92} & 39.42 & 39.54 & \underline{39.59} \\
HumanEval & 32.30 & 46.34 & 41.46 & 42.68 & \underline{48.78} & 43.90 & \textbf{49.39} & 48.17 \\
MBPP      & 36.60 & 37.80 & 31.20 & 31.40 & 37.80 & 34.80 & \underline{38.40} & \textbf{38.60} \\
\midrule
Avg       & 40.91 & 32.67 & 30.55 & 30.79 & 33.55 & 32.25 & \underline{33.83} & \textbf{34.20} \\
\bottomrule
\end{tabular}%
}
\end{table*}

\noindent \textbf{Cross-Architecture Distillation Is Effective.} Both \TIDE pipelines consistently outperform the non-distilled BD3LM baseline (with an average score of 32.67). The cross-tokenizer pipeline, utilizing the native \TIDECross strategy, achieves the highest average score of 34.20, while the shared-tokenizer pipeline reaches 33.55 using \TIDEShared. Furthermore, the baseline distillation objectives, even without the components of \TIDE, demonstrate improvement over the non-distilled model. This result confirms that the transfer of knowledge across architectures is viable across distinct tokenizers and attention mechanisms.

\noindent \textbf{Each Pipeline Favors Its Native Strategy.} The experimental results validate the modular design of \TIDE, as each pipeline benefits from a distinct optimal configuration. Within the cross-tokenizer pipeline, the native \TIDECross strategy outperforms the swapped \TIDEShared strategy by an average margin of 0.37. This observation indicates that the bounded gradients and dual-end noise filtering in the reverse objective are well-suited to scenarios involving alignment noise across different tokenizers (\secref{sec:objectives}). Conversely, within the shared-tokenizer pipeline, the native \TIDEShared strategy surpasses \TIDECross by an average margin of 2.76. This finding demonstrates that a progressive curriculum and enriched teacher signals are more effective when token-level alignment is exact.

\noindent \textbf{Distilled dLLMs Excel at Code Generation.} Across all configurations, the distilled models exhibit strong proficiency in programming tasks. On the HumanEval benchmark, \TIDEShared within the shared-tokenizer pipeline achieves a score of 48.78, and \TIDECross within the cross-tokenizer pipeline achieves 48.17 (\TIDEShared further reaches 49.39), all substantially exceeding the 32.30 score of an equivalent-sized autoregressive model. A similar pattern emerges on the MBPP benchmark, where the best distilled model reaches a score of 38.60, compared to 36.60 for the autoregressive baseline. This advantage suggests that the parallel generation process of diffusion decoding, which maintains global coherence across the entire output, is particularly suitable for structured outputs such as code. In these contexts, the syntactic and semantic consistency across the entire program remains critical.

\subsection{Ablation Studies}
\label{sec:ablation}

To isolate the contribution of each component within the \TIDECross (\TIDAL + \CompDemo) strategy, we conduct ablations on the shared-tokenizer pipeline (WeDLM $\to$ Qwen3-BD3LM) by removing one component at a time from the full method.
The three ablation conditions are: removing the timestep axis, replacing the dual-axis scheduling with a timestep-only schedule which serves as our baseline, and removing \CompDemo.
Detailed configuration settings and formal definitions for these ablations are provided in Appendix~\ref{app:details}.

\begin{table*}[!t]
\centering
\footnotesize
\setlength{\tabcolsep}{8pt}
\caption{Component-level ablation on the shared-tokenizer pipeline (WeDLM $\to$ Qwen3-BD3LM). \textbf{Bold}: best per row.}
\label{tab:component}
\begin{tabular}{l cccc}
\toprule
\textbf{Benchmark} 
& \textbf{Baseline} 
& \textbf{w/o Tstep} 
& \textbf{w/o \CompDemo} 
& \textbf{Full} \\
& (w/o Train) & & & (\TIDAL + \CompDemo) \\
\midrule
GSM8K     & 48.07 & 48.82 & \textbf{48.90} & \textbf{48.90} \\
MATH      & 11.74 & \textbf{11.96} & 11.84 & 11.68 \\
BBH       & 26.37 & 26.51 & \textbf{26.77} & 26.66 \\
MMLU-Pro  & 14.12 & \textbf{14.42} & 13.76 & 13.76 \\
HellaSwag & 40.03 & \textbf{40.35} & 40.16 & 40.27 \\
MMLU      & 39.81 & 39.84 & 39.58 & \textbf{39.92} \\
HumanEval & 45.73 & 43.90 & 44.51 & \textbf{46.95} \\
MBPP      & \textbf{38.60} & 37.20 & 38.20 & 37.00 \\
\midrule
Avg       & 33.06 & 32.88 & 32.97 & \textbf{33.14} \\
\bottomrule
\end{tabular}
\end{table*}

The complete method achieves the highest average score of 33.14 (Table~\ref{tab:component}), which confirms that each component provides a positive contribution.

\noindent \textbf{The Timestep Axis Is the Most Impactful Component.} The removal of the timestep axis causes the largest average performance drop of 0.26, which is primarily driven by a decline of 3.05 on HumanEval. This result validates the central motivation of \TIDAL: the reliability of the teacher varies according to the masking ratio, and the modulation of $\lamt$ along the diffusion timestep is essential for stable distillation. The timestep axis enables the student to decrease reliance on the teacher at high masking ratios where predictions are noisy, and to increase this reliance at low masking ratios where the teacher is confident. This dynamic is absent in autoregressive distillation, where the teacher always observes the complete left context.

\noindent \textbf{CompDemo Provides Consistent Gains.} The removal of \CompDemo reduces the average performance by 0.17, with the most notable drops occurring on HumanEval (2.44) and MMLU (0.34). The complementary mask splitting strategy enriches the guidance signal from the teacher by exposing the student to two complementary views per training sample, which proves particularly beneficial for tasks that require structured generation.

\noindent \textbf{Proposed Framework Outperforms Baseline.} Compared to the Baseline---which relies solely on timestep scheduling as proposed in prior works---our complete \TIDE framework achieves a higher average score. The integration of the training-progress axis into our dual-axis \TIDAL, combined with \CompDemo, stabilizes the early phase of distillation and prevents a significant performance drop of 0.83 on reasoning tasks such as GSM8K. This demonstrates the effectiveness and necessity of our proposed holistic approach over standalone scheduling methods.

\subsection{Inference Efficiency}
\label{sec:efficiency}

To evaluate the benefits of distillation for practical deployment, we benchmark inference efficiency under two settings on a single NVIDIA H100-80GB GPU in bfloat16 (Appendix~\ref{app:details}). The \emph{controlled} setting (Table~\ref{tab:efficiency}) generates exactly 256 tokens from a single fixed prompt for every model, providing a fair comparison of peak memory, latency, and throughput. The \emph{evaluation} setting (Table~\ref{tab:per_benchmark_speed}) reports throughput during actual benchmark runs, where input prompts, context lengths, and few-shot configurations vary across eight tasks.

\begin{table}[t]
\centering
\small
\caption{Inference efficiency comparison (controlled setting). Peak memory, latency, and throughput are measured on a single H100-80GB GPU generating 256 tokens in bfloat16.}
\label{tab:efficiency}
\begin{tabular}{lcccc}
\toprule
\textbf{Model} & \textbf{Params} & \textbf{Peak Mem} & \textbf{Latency} & \textbf{Tokens/s} \\
 & (B) & (GB) & (s) & \\
\midrule
\multicolumn{5}{l}{\emph{Student (BD3LM-0.6B)}} \\
\quad Distilled & 0.60 & 1.4 & 6.25 & 41.0 \\
\quad No Distill & 0.60 & 1.4 & 6.08 & 42.1 \\
\midrule
\multicolumn{5}{l}{\emph{AR Baseline}} \\
Qwen3-0.6B-Base & 0.60 & 1.2 & 4.99 & 51.3 \\
\midrule
\multicolumn{5}{l}{\emph{Teachers}} \\
WeDLM-8B-Instruct & 8.19 & 15.5 & 6.79 & 37.7 \\
LLaDA2.0-mini & 16.26 & 31.3 & 32.55 & 7.8 \\
\bottomrule
\end{tabular}
\end{table}

\begin{table}[t]
\centering
\small
\setlength{\tabcolsep}{3pt}
\caption{Per-benchmark inference speed (tokens/s) measured during actual evaluation runs. BD3LM exhibits near-constant throughput due to its fixed diffusion schedule.}
\label{tab:per_benchmark_speed}
\begin{tabular}{l ccc cc}
\toprule
& \multicolumn{2}{c}{\emph{Teachers}} & \emph{AR} & \multicolumn{2}{c}{\emph{Student}} \\
\cmidrule(lr){2-3} \cmidrule(lr){4-4} \cmidrule(lr){5-6}
\textbf{Benchmark}
& LLaDA2
& WeDLM
& Qwen3
& No Distill
& Distilled \\
\midrule
GSM8K     &  8.3 & 38.6 & 51.4 & 41.9 & 40.7 \\
MATH      & 10.3 & 38.4 & 51.1 & 41.8 & 40.6 \\
BBH       &  9.8 & 38.5 & 51.1 & 42.0 & 40.9 \\
MMLU-Pro  &  5.6 & 38.3 & 51.0 & 42.2 & 40.5 \\
HellaSwag &  6.8 & 37.2 & 51.5 & 42.4 & 40.6 \\
MMLU      &  6.2 & 37.1 & 51.4 & 41.9 & 41.1 \\
HumanEval &  9.8 & 36.8 & 51.1 & 42.5 & 41.7 \\
MBPP      &  9.1 & 37.0 & 51.8 & 42.3 & 41.6 \\
\midrule
Avg       &  8.2 & 37.7 & 51.3 & 42.1 & 40.9 \\
\bottomrule
\end{tabular}
\end{table}

\noindent \textbf{Distillation Enables Practical Deployment.} Under the controlled setting (Table~\ref{tab:efficiency}), the distilled student model requires only 1.4~GB of peak memory, representing a 22$\times$ reduction compared to LLaDA2 (31.3~GB) and an 11$\times$ reduction compared to WeDLM (15.5~GB). The inference latency of 6.25~s for 256 tokens yields a 5.2$\times$ speedup over LLaDA2 (32.55~s). As illustrated in \figref{fig:teaser}, cross-architecture distillation compresses the knowledge of large teacher \dLLMs into a model suitable for deployment on commodity hardware.

\noindent \textbf{Distillation Adds Minimal Overhead.} Under the controlled setting, distillation introduces only a 2.6\% reduction in throughput relative to the undistilled BD3LM (41.0 vs.\ 42.1~tokens/s), with a marginal latency increase (6.25~s vs.\ 6.08~s) and identical memory footprint. The evaluation setting (Table~\ref{tab:per_benchmark_speed}) confirms that this overhead is uniform across all eight benchmarks despite varying prompt lengths and generation requirements, indicating that the distillation procedure does not degrade inference efficiency under realistic conditions. Compared to the AR baseline of the same size (51.3~tokens/s), the BD3LM student achieves approximately 80\% throughput due to the iterative diffusion process; however, the quality gains from distillation (Table~\ref{tab:main_combined}) and the inherent advantages of block-parallel generation make this trade-off favorable for practical deployment.

\section{Conclusion}
\label{sec:conclusion}

We present \TIDE, the first cross-architecture distillation framework for heterogeneous dLLMs. Experiments across two pipelines and eight benchmarks show that (1)~cross-architecture distillation improves the baseline by +1.53 on average, (2)~each pipeline favors a distinct strategy (Reverse \CALM for cross-tokenizer, \TIDAL + \CompDemo for shared-tokenizer), and (3)~distilled dLLMs outperform the same-size AR model by +16.48 on HumanEval. Future work includes scaling student capacity and extending the framework to continuous-state diffusion LMs.

\section*{Ethics Statement}
This work focuses on improving the efficiency of diffusion language models through knowledge distillation. Publicly available datasets and pre-trained models are utilized. 
The computational requirements are moderate. No direct negative societal impacts are foreseen beyond the general concerns associated with the deployment of language models.

\section*{LLM Usage}
In this section, we clarify the role of large language models (LLMs) in preparing this work. 
The model was used exclusively for language polishing, such as refining grammar, style, and readability, without contributing to the research design, analysis, or conclusions.

\section*{Acknowledgments}
We would like to sincerely thank Xiangtai Li and Anran Wang for their selfless guidance and invaluable support throughout this project.

\bibliography{colm2026_conference}
\bibliographystyle{colm2026_conference}

\clearpage
\appendix

\section{Related Work}
\label{app:related}

\noindent
\textbf{Diffusion Language Models.}
D3PM~\citep{austin2021d3pm} pioneered discrete diffusion for text generation. MDLM~\citep{sahoo2024mdlm} and SEDD~\citep{lou2024sedd} further established theoretical foundations through simplified masked diffusion and score estimation, respectively.
Building on these foundations, practical dLLMs have emerged with diverse architectures: LLaDA~\citep{nie2025llada} adopts full bidirectional attention, BD3LM~\citep{arriola2025bd3lm} introduces block diffusion with staircase attention, Dream~\citep{ye2025dream} extends masked diffusion with rectified estimation, and WeDLM~\citep{liu2025wedlm} proposes a causal diffusion architecture combining sliding-window and global attention. DiffuLLaMA~\citep{gong2025diffullama} converts pre-trained AR models into diffusion LMs.
This architectural heterogeneity---encoder, decoder-block, and causal variants---motivates the need for a cross-architecture distillation framework.

\noindent
\textbf{Knowledge Distillation of Large Language Models.}
Knowledge distillation~\citep{hinton2015distilling} transfers knowledge from a large teacher to a smaller student through soft targets. For AR models, representative methods include reverse KL minimization~\citep{gu2024minillm}, skewed KL divergence~\citep{ko2024distillm}, on-policy distillation~\citep{agarwal2024gkd}, dual-space transfer~\citep{zhang2024dskd}, feature-level distillation~\citep{saadi2025flexkd}, and time-varying interpolation~\citep{taid2025}.
These methods target the left-to-right AR paradigm. \TIDE builds on this line of work, particularly TAID's interpolation principle, and adapts it to dLLMs, where teacher reliability varies with the diffusion timestep and all masked tokens are predicted simultaneously.

\noindent
\textbf{Distillation for Diffusion Language Models.}
Existing dLLM distillation methods---CDLM~\citep{kim2025cdlm}, DDD~\citep{hayakawa2024ddd}, LSD~\citep{fu2025lsd}, and SDTT~\citep{sdtt2025}---focus exclusively on \emph{step compression}: the student and teacher share the same architecture and tokenizer, and the goal is to reduce the number of inference steps.
\TIDE addresses a fundamentally different problem: \emph{cross-architecture} distillation, where the teacher and student differ in architecture, attention mechanism, and potentially tokenizer. For the cross-tokenizer case, we build on ALM~\citep{minixhofer2025alm} and ZeTT~\citep{minixhofer2024zett}, adapting chunk-level approximate likelihood matching from the AR setting to dLLMs as \CALM (\secref{sec:alm}).

\section{Training, Inference, and Evaluation Details}
\label{app:details}

This section provides the comprehensive configurations for training, inference, and evaluation used throughout the experiments.

\noindent \textbf{Training Configurations.}

Table~\ref{tab:hyperparams} summarizes the complete set of training hyperparameters for both pipelines.

\begin{table}[!h]
\centering
\small
\caption{Training hyperparameters.}
\label{tab:hyperparams}
\begin{tabular}{lcc}
\toprule
\textbf{Parameter} & \textbf{Cross-Tok.} & \textbf{Shared-Tok.} \\
\midrule
Teacher & LLaDA2.0-mini (16B MoE) & WeDLM-8B-Instruct (8B) \\
Student init & SFT v0.1 & SFT v0.1 \\
Methods & Reverse \CALM & \TIDAL + \CompDemo \\
Learning rate & 5e-5 & 5e-5 \\
Epochs & 10 & 10 \\
Sequence length (student) & 512 & 512 \\
Teacher max length & 1024 & 768 \\
Block size & 32 & 32 \\
Precision & bfloat16 & bfloat16 \\
Teacher precision & bfloat16 & bfloat16 \\
Attention impl. & flex\_attention & flex\_attention \\
Strategy & DDP & DDP \\
Dataset & \multicolumn{2}{c}{Tulu-3 SFT + SmolTalk + OpenCoder-SFT (Python)} \\
\TIDAL $\laminit \to \lammax$ & - & $0.1 \to 0.9$ \\
\TIDAL schedule & - & cosine \\
\TIDAL timestep weight & - & midrange \\
\CompDemo demo\_ratio & - & 0.5 \\
Temperature $T$ & - & 2.0 \\
\bottomrule
\end{tabular}
\end{table}

\noindent \textbf{Ablation Study Configurations.}
For the component-level ablation studies presented in Section~\ref{sec:ablation}, all configurations are trained for 3 epochs. The specific definitions for the three ablation conditions are as follows:
\begin{itemize}[leftmargin=*,nosep]
    \item \textbf{w/o Tstep (removing the timestep axis):} $\lamt = \lamtrain$, \ie lambda depends only on the training progress.
    \item \textbf{Baseline (timestep-only schedule):} $\lamt = \mathrm{const} \times (1-t)$, replacing the dual-axis scheduling with a schedule proposed in previous works.
    \item \textbf{w/o \CompDemo:} Removes the complementary demonstration strategy, using only a single teacher forward pass with the dual-axis \TIDAL.
\end{itemize}

\noindent \textbf{Evaluation Protocol.}
All evaluations employ diffusion sampling with a block size of 32 and a classifier-free guidance scale of 0.0.
The number of sampling steps varies depending on the specific task, ranging from 3 to 256. Table~\ref{tab:eval_config} details the precise configuration for each evaluation.

\begin{table}[!h]
\centering
\small
\caption{Per-task evaluation configuration.}
\label{tab:eval_config}
\begin{tabular}{lcccc}
\toprule
\textbf{Task} & \textbf{max\_new\_tokens} & \textbf{steps} & \textbf{few-shot} & \textbf{batch\_size} \\
\midrule
MMLU & 3 & 3 & 0 & 128 \\
MMLU-Pro & 256 & 256 & 0 & 64 \\
HellaSwag & 3 & 3 & 0 & 128 \\
GSM8K & 256 & 256 & 0 & 128 \\
BBH & 256 & 256 & 3 & 32 \\
MATH & 256 & 256 & 0 & 32 \\
HumanEval & 256 & 256 & 0 & 128 \\
MBPP & 256 & 256 & 0 & 32 \\
\bottomrule
\end{tabular}
\end{table}

\noindent \textbf{Inference Efficiency Protocol.}
For the inference efficiency measurements in Section~\ref{sec:efficiency}, all evaluations are conducted on a single NVIDIA H100-80GB GPU with bfloat16 precision. Under the controlled setting, we generate 256 tokens and report the best performance observed across five independent runs. Under the evaluation setting, we run 50 randomly sampled examples from each benchmark and report the average.

\section{Gradient Analysis}
\label{app:gradient}

This appendix provides detailed derivations of the gradients that support the theoretical justification for Reverse \CALM presented in Section~\ref{sec:objectives}.

\noindent \textbf{Forward CALM gradient.} The gradient of the forward CALM loss with respect to the student parameters $\theta$ contains a ratio $p_t^c / p_s^c$, where $p_t^c$ and $p_s^c$ are the chunk probabilities of the teacher and the student, respectively:
\begin{equation}
g_{\text{fwd}} = \frac{p_t^c}{p_s^c} - \frac{1 - p_t^c}{1 - p_s^c}.
\end{equation}
When $p_s^c \to 0$ but $p_t^c > 0$, $g_{\text{fwd}} \to +\infty$. In the cross-tokenizer setting, imperfect chunk alignment frequently produces chunks where the student assigns a low initial probability, triggering this gradient explosion.

\noindent \textbf{Reverse CALM gradient.} The reverse CALM gradient takes the form:
\begin{equation}
\frac{\partial \Ls_{\text{rev}}}{\partial \theta} = -\sum_c \frac{\partial p_s^c}{\partial \theta} \cdot \log \frac{p_t^c}{1 - p_t^c}.
\end{equation}
The gradient coefficient $\log \frac{p_t^c}{1 - p_t^c}$ depends solely on the fixed teacher probabilities and is bounded: $|g_{\text{rev}}| \leq |\log \frac{1-\epsilon}{\epsilon}|$ for $p_t^c \in [\epsilon, 1-\epsilon]$. This provides a stable, self-selecting training signal where the student naturally concentrates updates on the high-probability modes of the teacher.

\noindent \textbf{Dual-end noise filtering.} The full reverse gradient $\frac{\partial p_s^c}{\partial \theta} \cdot \log \frac{p_t^c}{1-p_t^c}$ is filtered on both ends: (1)~poorly aligned chunks yield $p_t^c \approx 0.5$, zeroing the teacher-end coefficient; (2)~chunks with low $p_s^c$ contribute small $\frac{\partial p_s^c}{\partial \theta}$, suppressing the student-end signal. Forward CALM has no such dual filtering---$1/p_s^c$ amplifies noise instead.

\noindent \textbf{Bernoulli KL equivalence.} Reverse CALM is equivalent to minimizing the Bernoulli KL divergence $\text{KL}_{\text{Bern}}(p_s^c \| p_t^c) = p_s^c \log \frac{p_s^c}{p_t^c} + (1-p_s^c) \log \frac{1-p_s^c}{1-p_t^c}$, up to an additive constant independent of $\theta$. This gives Reverse CALM a principled information-theoretic interpretation as mode-seeking distillation in Bernoulli (scalar) space.

\noindent \textbf{TIDAL and reverse direction.} \TIDAL addresses the instability of forward CALM via a curriculum where $\lamt$ transitions from emphasizing the student to the teacher. However, reverse CALM does not exhibit this instability. Applying \TIDAL to reverse CALM is counterproductive: during the late stages of training, the $(1-\lamt)$ factor approaches 0.1, suppressing the gradient and destroying the self-selection mechanism of reverse CALM. This confirms that \TIDAL is effective with forward-direction objectives but should not be applied to reverse-direction objectives.

\section{Limitations and Future Work}
\label{app:limitations_and_future}

The empirical scope of this work is limited to a 0.6B-parameter student model using block diffusion with staircase attention. A primary avenue for future research is to scale the student model to 1.3B or 3B parameters to assess whether cross-architecture distillation efficiency improves as the capacity gap narrows. Furthermore, while the \TIDE framework is theoretically architecture-agnostic, empirical validation on alternative structures, such as continuous-state diffusion language models or encoder-style \dLLMs, remains necessary. Adapting the proposed loss formulations from categorical distributions to continuous densities is a critical next step to broaden the framework's applicability.

Furthermore, the training pipeline currently operates within a 512-token context window, leaving the efficacy of cross-tokenizer alignment and \CompDemo on extended sequences unexplored. Future investigations must examine how an increase in the number of alignment chunks alters the relative contributions of these components. 
Additionally, the present methodology processes the cross-tokenizer and shared-tokenizer pipelines independently; formulating a unified multi-teacher distillation objective could facilitate complementary knowledge transfer and yield a more robust representation for the student model.

Finally, computational efficiency and optimization dynamics present crucial areas for refinement. The \CompDemo component necessitates two forward passes through the frozen teacher model per step, increasing training duration by approximately 50\%. Concurrently, gradient-suppression mechanisms render the combination of Reverse \CALM and \TIDAL counterproductive (Appendix~\ref{app:gradient}). Developing alternative scheduling paradigms, such as restricting \ TIDAL's modulation exclusively to the cross-entropy objective, is essential for reconciling these optimization conflicts and realizing the cumulative benefits of both strategies.

\section{Case Study}
\label{app:case_study}

To comprehend the information conveyed by knowledge distillation beyond aggregate benchmark improvements, we conduct a diagnostic study that examines the transfer of dark knowledge and qualitative error patterns.

\begin{figure}[h]
\centering
\includegraphics[width=0.55\columnwidth]{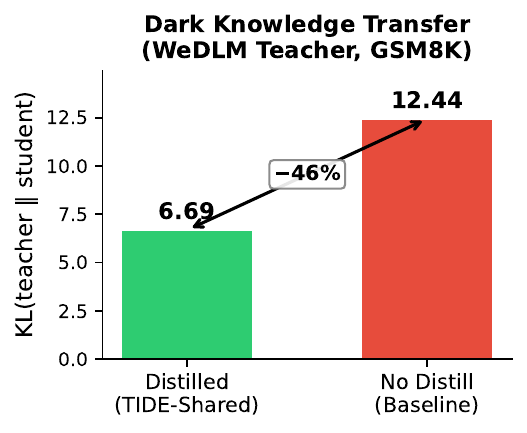}
\caption{The KL divergence relative to the WeDLM teacher on the GSM8K dataset. The distilled student achieves a KL divergence that is 46\% lower than that of the non-distilled baseline ($6.69$ compared to $12.44$).}
\label{fig:kl_wedlm}
\end{figure}

\noindent \textbf{Dark Knowledge Transfer.} Within the shared-tokenizer pipeline, we measure the KL divergence between the predictions of the student and those of the teacher at intermediate denoising timesteps. As illustrated in Figure~\ref{fig:kl_wedlm}, \TIDEShared reduces the KL divergence relative to the WeDLM teacher by 46\% on the GSM8K dataset ($6.69$ compared to $12.44$). This reduction confirms that the distilled student inherits the teacher's prediction distribution. 
The cross-tokenizer KL comparison (from LLaDA2 to the student) is omitted due to vocabulary misalignment.

\noindent \textbf{Qualitative Error Analysis.} We select four instances wherein the non-distilled baseline fails, whereas exactly one distilled variant succeeds (Table~\ref{tab:case_studies}). Table~\ref{tab:case_study_details} illustrates two representative examples in detail. For instance, \#39 of GSM8K, the baseline confuses time allocation with speed in a multi-step distance problem; only \TIDECross (distilled from LLaDA2) correctly computes the travel distances. For instance, \#50 of HumanEval, the baseline calls \texttt{encode\_shift} again instead of reversing the shift; only \TIDEShared (distilled from WeDLM) correctly subtracts~5. These examples demonstrate that each teacher imparts distinct knowledge. Specifically, LLaDA2 corrects multi-step arithmetic omissions and enhances type-level awareness in code generation, whereas WeDLM resolves proportional reasoning errors and facilitates algorithmic inversion.

\begin{table*}[p]
\centering
\small
\caption{A detailed textual representation of case studies. This table compares the baseline model's reasoning chains and generated code against those of distilled variants.}
\label{tab:case_study_details}
\begin{tabular}{@{} p{0.18\textwidth} p{0.52\textwidth} p{0.25\textwidth} @{}}
\toprule
\multicolumn{3}{@{}l@{}}{\textbf{GSM8K \#39: Multi-Step Distance Calculation}} \\
\multicolumn{3}{@{}p{0.95\textwidth}@{}}{\textit{Prompt:} Dana can run at a rate of speed four times faster than she can walk, but she can skip at a rate of speed that is half as fast as she can run. If she can skip at 3 miles per hour, how many miles can she travel in six hours if she spends one-third of the time running and two-thirds of the time walking? (\textit{Ground Truth:} 18)} \\
\midrule
\textbf{Model} & \textbf{Output} & \textbf{Analysis} \\
\midrule
\textbf{Baseline (No Distill)} &
{\footnotesize Dana can skip at a rate of speed 3 miles per hour, so she can run at a rate of speed $3 * 2 = 6$ miles per hour. If she spends one-third of the time running, she will run $6 / 3 = 2$ miles per hour. If she spends two-thirds of the time walking, she will walk $6 / 3 = 2$ miles per hour. The total distance Dana can travel in 6 hours is $2 * 2 + 4 * 2 = 12 + 4 = 16$ miles. Answer: \textbf{16}} &
Confuses time allocation ($6/3\!=\!2$ hours) with speed (``2 miles per hour''). \\
\midrule
\textbf{\TIDECross (LLaDA2)} &
{\footnotesize Dana can skip at a rate of 3 miles per hour, so she can run at a rate of $3 * 2 = 6$ miles per hour. If she spends one-third of the time running, she will run for $6 / 3 = 2$ hours. If she spends two-thirds of the time walking, she will walk for $6 / 3 = 2$ hours. In 2 hours, she will run for $2 * 6 = 12$ miles. In 2 hours, she will walk for $2 * 3 = 6$ miles. Therefore, Dana can travel a total of $12 + 6 = 18$ miles in six hours. Answer: \textbf{18}} &
Correctly computes time allocation and travel distance. \\
\midrule
\textbf{\TIDEShared (WeDLM)} &
{\footnotesize Dana can skip at a rate of 3 miles per hour, so she can run at a rate of $3 * 2 = 6$ miles per hour. If she spends one-third of the time running, she will spend $6 / 3 = 2$ hours running. If she spends two-thirds of the time walking, she will spend $6 / 3 = 2$ hours walking. In total, she will spend $2 + 2 = 4$ hours running and walking. If she can run at a rate of 6 miles per hour, she will travel $6 * 6 = 36$ miles in 6 hours of running. If she can skip at a rate of 3 miles per hour, she will travel $3 * 6 = 18$ miles in 6 hours skipping. Therefore, she will travel a total of $36 + 18 = 54$ miles. Answer: \textbf{54}} &
Correctly allocates time but uses the full 6 hours instead of the allocated 2 hours. \\
\bottomrule
\toprule
\multicolumn{3}{@{}l@{}}{\textbf{HumanEval \#50: Algorithmic Inversion}} \\
\multicolumn{3}{@{}p{0.95\textwidth}@{}}{\textit{Prompt:} The function \texttt{encode\_shift} shifts each character by +5 in the alphabet. Implement the function \texttt{decode\_shift}: reverse the encoding, and return the original string.} \\
\midrule
\textbf{Model} & \textbf{Output} & \textbf{Analysis} \\
\midrule
\textbf{Baseline (No Distill)} &
{\footnotesize\ttfamily
\begin{tabular}[t]{@{}l@{}}
return encode\_shift(s)
\end{tabular}} &
Calls \texttt{encode\_shift} again (+5) instead of reversing ($-$5). \\
\midrule
\textbf{\TIDECross (LLaDA2)} &
{\footnotesize\ttfamily
\begin{tabular}[t]{@{}l@{}}
return encode\_shift(s)
\end{tabular}}

{\vspace{1mm}\footnotesize\textrm{+ 500-char explanation that incorrectly justifies this as valid decoding.}} &
Same erroneous code as Baseline, but with a longer (incorrect) explanation appended. \\
\midrule
\textbf{\TIDEShared (WeDLM)} &
{\footnotesize\ttfamily
\begin{tabular}[t]{@{}l@{}}
return "".join( \\
~~[chr(((ord(ch) - 5 \\
~~~~- ord("a")) \% 26) \\
~~~~+ ord("a")) \\
~~for ch in s])
\end{tabular}} &
Correctly subtracts 5 to reverse the shift. \\
\bottomrule
\end{tabular}
\end{table*}

\begin{table}[h]
\centering
\small
\caption{A summary of the case studies. The non-distilled baseline fails on all four problems, whereas each distillation pipeline resolves distinct errors.}
\label{tab:case_studies}
\begin{tabular}{p{4.2cm}ccc}
\toprule
\textbf{Case \& Error Type} & \textbf{Base} & \textbf{\TIDECross} & \textbf{\TIDEShared} \\
\midrule
\multicolumn{4}{l}{\emph{Math Reasoning (GSM8K)}} \\
\#39: rate-distance confusion in multi-step problem & $\times$ & $\checkmark$ & $\times$ \\
\#77: proportional relationships reversed & $\times$ & $\times$ & $\checkmark$ \\
\midrule
\multicolumn{4}{l}{\emph{Code Generation (HumanEval)}} \\
\#11: bitwise XOR on string characters & $\times$ & $\checkmark$ & $\times$ \\
\#50: decode calls encode (no reversal) & $\times$ & $\times$ & $\checkmark$ \\
\bottomrule
\end{tabular}
\end{table}

\subsection*{Verbatim Model Outputs}

The following presents the complete, unedited model outputs for all four cases in Table~\ref{tab:case_studies}. These raw responses confirm that each model produces a distinct reasoning chain or code implementation.

\paragraph{GSM8K \#39: Baseline (No Distill) --- Incorrect.}
\begin{quote}\small
Dana can skip at a rate of speed 3 miles per hour, so she can run at a rate of speed $3 \times 2 = 6$ miles per hour. If she spends one-third of the time running, she will run $6 / 3 = 2$ miles per hour. If she spends two-thirds of the time walking, she will walk $6 / 3 = 2$ miles per hour. The total distance Dana can travel in 6 hours is $2 \times 2 + 4 \times 2 = 12 + 4 = 16$ miles. \#\#\#\# 16. The answer is: 16
\end{quote}

\paragraph{GSM8K \#39: \TIDECross (LLaDA2) --- Correct.}
\begin{quote}\small
Dana can skip at a rate of 3 miles per hour, so she can run at a rate of $3 \times 2 = 6$ miles per hour. If she spends one-third of the time running, she will run for $6 / 3 = 2$ hours. If she spends two-thirds of the time walking, she will walk for $6 / 3 = 2$ hours. In 2 hours, she will run for $2 \times 6 = 12$ miles. In 2 hours, she will walk for $2 \times 3 = 6$ miles. Therefore, Dana can travel a total of $12 + 6 = 18$ miles in six hours. \#\#\#\# 18. The answer is: 18
\end{quote}

\paragraph{GSM8K \#39: \TIDEShared (WeDLM) --- Incorrect.}
\begin{quote}\small
Dana can skip at a rate of 3 miles per hour, so she can run at a rate of $3 \times 2 = 6$ miles per hour. If she spends one-third of the time running, she will spend $6 / 3 = 2$ hours running. If she spends two-thirds of the time walking, she will spend $6 / 3 = 2$ hours walking. In total, she will spend $2 + 2 = 4$ hours running and walking. If she can run at a rate of 6 miles per hour, she will travel $6 \times 6 = 36$ miles in 6 hours of running. If she can skip at a rate of 3 miles per hour, she will travel $3 \times 6 = 18$ miles in 6 hours skipping. Therefore, she will travel a total of $36 + 18 = 54$ miles in 6 hours. \#\#\#\# 54. The answer is: 54
\end{quote}

\paragraph{GSM8K \#77: Baseline (No Distill) --- Incorrect.}
\begin{quote}\small
Sarah does 400 pounds of laundry. Raymond does $4 \times 400 = 1600$ pounds of laundry. Raymond does $1600 / 2 = 800$ pounds of laundry. Raymond does $800 - 400 = 400$ pounds of laundry than David.
\end{quote}

\paragraph{GSM8K \#77: \TIDECross (LLaDA2) --- Incorrect.}
\begin{quote}\small
$400 / 4 = 100$ pounds of laundry. Raymond does $100 / 2 = 50$ pounds of laundry. The difference is $100 - 50 = 50$ pounds.
\end{quote}

\paragraph{GSM8K \#77: \TIDEShared (WeDLM) --- Correct.}
\begin{quote}\small
Sarah does 400 pounds of laundry. Sarah does 4 times as much laundry as David, so David does $400 / 4 = 100$ pounds of laundry. Raymond does half as much laundry as Sarah, so Raymond does $400 / 2 = 200$ pounds of laundry. The difference in the amount of laundry Raymond and David do is $200 - 100 = 100$ pounds.
\end{quote}

\paragraph{HumanEval \#11 (\texttt{string\_xor}): Baseline (No Distill) --- Failed.}
{\small
\begin{verbatim}
    result = ""
    for a, b in zip(a, b):
        result += str(a ^ b)
    return result
\end{verbatim}
}

\paragraph{HumanEval \#11 (\texttt{string\_xor}): \TIDECross (LLaDA2) --- Passed.}
{\small
\begin{verbatim}
    result = ''
    for a, b in zip(a, b):
        if a != b:
            result += '1'
        else:
            result += '0'
    return result
\end{verbatim}
}

\paragraph{HumanEval \#11 (\texttt{string\_xor}): \TIDEShared (WeDLM) --- Failed.}
{\small
\begin{verbatim}
    result = ""
    for a, b in zip(a, b):
        result += str(a ^ b))
    return result[:-1]
\end{verbatim}
}

\paragraph{HumanEval \#50 (\texttt{decode\_shift}): Baseline (No Distill) --- Failed.}
{\small
\begin{verbatim}
    return encode_shift(s)
\end{verbatim}
}

\paragraph{HumanEval \#50 (\texttt{decode\_shift}): \TIDECross (LLaDA2) --- Failed.}
{\small
\begin{verbatim}
    return encode_shift(s)
\end{verbatim}
}
\noindent\textit{Note: The generated code is identical to the Baseline. \TIDECross additionally provides a lengthy explanation that incorrectly characterizes this as a valid decoding strategy, suggesting that the model lacks an understanding that decoding requires a reversal operation.}

\paragraph{HumanEval \#50 (\texttt{decode\_shift}): \TIDEShared (WeDLM) --- Passed.}
{\small
\begin{verbatim}
    return "".join(
        [chr(((ord(ch) - 5 - ord("a"))
              % 26) + ord("a"))
         for ch in s])
\end{verbatim}
}

\end{document}